\documentclass{article}

\usepackage{lineno}
\usepackage{booktabs}
\usepackage{authblk}
\usepackage{multirow}
\usepackage{graphicx}
\usepackage[numbers]{natbib}
\usepackage{hyperref}
\usepackage{fixltx2e}
\usepackage{subcaption}
\usepackage{amsmath}
\usepackage{lscape}
\usepackage[preprint]{tackling_climate_workshop_style}

\usepackage[utf8]{inputenc} % allow utf-8 input
\usepackage[T1]{fontenc}    % use 8-bit T1 fonts
\usepackage{hyperref}       % hyperlinks
\usepackage{url}            % simple URL typesetting
\usepackage{booktabs}       % professional-quality tables
\usepackage{amsfonts}       % blackboard math symbols
\usepackage{nicefrac}       % compact symbols for 1/2, etc.
\usepackage{microtype}      % microtypography

\title{\textit{} Domain-Adaptive Climate Downscaling Under Temporal Distribution Shift}

\author[1]{\textbf{Shuochen Wang}}
\author[2]{\textbf{Nishant Yadav}}
\author[1, 3]{\textbf{Auroop R. Ganguly}}
\affil[1]{Sustainability and Data Sciences Laboratory, Northeastern University}
\affil[2]{Microsoft}
\affil[3]{AI for Climate and Sustainability, The Institute for Experiential AI, Northeastern University}

\begin{document}

\maketitle

\begin{abstract}
Deep-learning-based climate downscaling aims to learn relationships from historical low-resolution (LR) and high-resolution (HR) climate data to generate HR climate projections. However, this setting faces a temporal out-of-distribution (OOD) challenge: models trained on historical data are commonly applied to future projections whose distributions may differ substantially from the training period. This study investigates temporal OOD shift for daily temperature downscaling over the Continental United States using paired LR-HR model simulations. We propose a temporal domain-adaptive downscaling framework that combines supervised HR reconstruction on historical data with domain alignment between historical and future climate distributions. Experiments across future validation periods show that the proposed domain-adaptive model consistently outperforms statistical and deep-learning-based bias-correction methods, with the largest gains occurring when the temporal distribution shift is strongest. Spatial analyses indicate stronger improvements over high-elevation and topographically complex regions, along with higher spatiotemporal correlation with the HR target. The extreme analysis shows that domain adaptation also reduces upper-tail temperature bias relative to the non-adaptive model. These results demonstrate that temporal domain adaptation can improve the robustness of HR climate projections under non-stationary climate conditions.
\end{abstract}

\section{Introduction}
Global Climate Models (GCMs) are essential tools for understanding the Earth system, simulating historical climate variability, and projecting future climate change under different forcing scenarios. Despite advances in Earth's system representation, GCMs from the current generation Coupled Model Intercomparison Project (CMIP) remain too coarse for many regional and local applications, with horizontal grid spacings often on the order of tens to hundreds of kilometers \cite{eyring2016overview,chen2021framing}. At these resolutions, fine-scale effects associated with topography, coastlines, land-sea contrasts, and unresolved sub-grid-scale physical processes cannot be explicitly represented and must instead be approximated through parameterizations, which can introduce biases and smooth local climate variability \cite{chen2021framing,rasp2018deep}. Downscaling methods have therefore been developed to bridge the scale gap between low-resolution (LR) GCM outputs and the high-resolution (HR) climate information needed for regional impact assessment \cite{houghton2001climate,wilby2004guidelines}. These methods are commonly grouped into dynamical and statistical approaches. Dynamical downscaling uses Regional Climate Models (RCMs) to refine global model outputs through HR physical simulations \cite{giorgi1991approaches,giorgi2015regional,maraun2010precipitation}. A major advantage of RCMs is that they retain a process-based representation of regional climate dynamics, including interactions with terrain, land-surface properties, and coastlines \cite{adachi2020methodology}. However, RCMs are computationally expensive and difficult to scale across large ensembles, emission scenarios, long simulation periods, and complex model configurations. Their projections are also affected by multiple uncertainty sources, including internal climate variability, scenario uncertainty, and systematic errors inherited from the driving GCMs \cite{adachi2020methodology,hawkins2009potential}. Statistical downscaling, by contrast, infers empirical relationships between coarse-scale predictors and fine-scale climate observations or simulations \cite{wilby2004guidelines}. Classical methods include regression-based models \cite{wilby1997downscaling}, canonical correlation analysis \cite{von1993downscaling}, analog methods \cite{zorita1999analog,maurer2008utility}, and bias correction or bias-correction spatial disaggregation approaches \cite{maurer2008utility,cannon2015bias,gudmundsson2012downscaling}. These methods are computationally efficient, but their reliability is limited by stationarity assumptions, sensitivity to predictor selection and observational quality, difficulty preserving spatial and temporal dependence, and weak representation of extremes \cite{wilby1997downscaling,hewitson2014interrogating,maraun2013bias,hempel2013trend}. These limitations motivate data-driven downscaling approaches that can learn fine-scale mappings from coarse climate fields while incorporating HR geographic information.

Machine learning (ML) extends statistical downscaling by learning nonlinear, spatially structured mappings from LR climate predictors to HR target fields using paired observations and simulations. Many recent approaches formulate downscaling as a super-resolution problem in deep learning (DL) and computer vision, in which a coarse climate field is treated analogously to a LR image and the model learns to reconstruct the missing fine-scale structure of the corresponding HR field \cite{vandal2017deepsd,rodrigues2018deepdownscale,bano2020configuration}. In a typical super-resolution architecture, LR inputs are encoded through feature-extraction layers, transformed by nonlinear mapping blocks such as residual convolution, and then upsampled onto the HR grid. Depending on the target variable, models can utilize LR temperature, precipitation, wind, pressure, and other atmospheric fields, together with HR auxiliary variables, such as elevation and land-sea masks, to guide the reconstruction of terrain-driven gradients, coastal effects, and other local heterogeneity \cite{vandal2017deepsd,liu2020climate,wang2021deep,vaughan2021convolutional}. More recent innovations adapt advanced image super-resolution architectures, including transformer-based models and generative models such as conditional Generative Adversarial Networks (GANs) and diffusion models, to better capture multiscale spatial dependencies, sharpen fine-scale patterns, and represent stochastic uncertainty in unresolved sub-grid variability \cite{schmidt2025generative,aich2026conditional,watt2024generative,zhong2024investigating}.

A key remaining challenge for DL-based downscaling is temporal out-of-distribution (OOD) shift under a changing climate. Most downscaling models are trained on historical LR-HR pairs and then applied to future climate projections, implicitly assuming that the learned relationship between coarse predictors and fine-scale targets remains valid outside the historical training period. However, anthropogenic warming can shift the marginal distributions of predictors and targets, alter the frequency of extremes, and create no-analog climate states, making future projections a non-stationary target domain rather than a simple extension of the training distribution \cite{rampal2024enhancing, lanzante2018some,hamed2026machine}. Recent studies have begun to examine this problem directly, showing that DL models can produce plausible mean climate change signals in some settings but may still suffer from extrapolation artifacts, poor transferability across GCMs, and large errors for future extremes \cite{bano2021suitability,hernanz2024limitations}. Several approaches have been proposed to improve robustness, including trend-preserving bias correction combined with DL, physically constrained downscaling, probabilistic generative models, and multi-dataset pretraining to improve transferability \cite{abdelmoaty2026does,quesada2022repeatable,doury2023regional,wang2024multivariate,prasad2024evaluating}.

A promising direction is to frame historical-to-future downscaling as a domain adaptation problem, a branch of transfer learning designed for settings where a model trained on one data distribution must generalize to another. Domain adaptation has been widely studied in computer vision, where ML models trained on a labeled source domain are adapted to a target domain, whose distribution differs from the training data. Instead of relying only on supervised fine-tuning, domain adaptation methods explicitly reduce the discrepancy between source and target feature distributions. Common strategies include moment-matching losses such as Maximum Mean Discrepancy (MMD) \cite{tzeng2014deep} and correlation alignment \cite{sun2016deep}, which align source and target feature statistics, and adversarial approaches such as Domain Adversarial Neural Networks (DANN) \cite{ganin2016domain}, where a domain classifier encourages the encoder to learn features that are useful for the prediction task but insensitive to domain identity \cite{tzeng2017adversarial}. Beyond conventional computer vision, domain adaptation has also been applied in Earth observation tasks, including land-cover classification and crop mapping \cite{martini2021domain,nyborg2022timematch,wang2022cross,capliez2023multisensor}. This framework is well suited to climate downscaling because historical observations and simulations can be treated as a labeled source domain, while future projections represent a temporally shifted target domain with different thermodynamic conditions. In practice, a downscaling model can be trained on labeled historical LR-HR pairs while using unlabeled future LR fields to align source and target representations through domain adaptation. For temporal OOD downscaling, domain adaptation offers a way to reduce historical-future distribution shift while preserving the physically meaningful mapping from large-scale climate predictors to fine-scale regional structure.

In this study, we explore domain-adaptive DL for climate downscaling under temporal distribution shift. To address the non-stationarity between historical and future climate conditions, we investigate whether domain adaptation can improve the transferability of DL-based downscaling models beyond the historical training period. By encouraging the model to learn representations that are less sensitive to temporal domain shifts, this approach aims to support more reliable HR climate projections under changing climate conditions.

\section{Data and Preprocessing}
This study uses daily near-surface air temperature from three paired GCM-RCM simulations: CanESM2-RCA4 \cite{arora2011carbon,samuelsson2015surface}, CanESM2-CanRCM4 \cite{scinocca2016coordinated}, and EC-EARTH-RCA4 \cite{hazeleger2012ec}. We select these three pairs to assess the stability of the proposed methods across datasets with different biases between the RCM and its driving GCM. For each pair, the LR GCM field is used as the predictor, and the corresponding HR RCM field is used as the downscaling target. Unlike a perfect-model framework, in which LR inputs are derived directly from the HR target, this GCM-RCM configuration forms an imperfect-model framework. The downscaling model must therefore account for biases between the paired datasets, making this a more challenging downscaling task \cite{rampal2024enhancing,doury2023regional,fallah2025climate}.

The GCM data were obtained from the Earth System Grid Federation (ESGF) \cite{cinquini2014earth}. The RCM data were obtained from the Coordinated Regional Climate Downscaling Experiment (CORDEX) North American NAM-44i domain \cite{ncar_gdex_dataset_d316009}. We use the raw CORDEX outputs without pre-applied spatial or temporal bias correction. The study region is centered on the Contiguous United States (CONUS), bounded by 24--50°N and 125--66°W. Both GCM and RCM fields are first cropped to the study region. Raw CORDEX RCM outputs on the native NAM-44i rotated-pole grid, approximately 0.44° or 50 km, are regridded to a regular 0.5° latitude-longitude grid using bilinear interpolation. The GCM fields are interpolated from their native model grids to a common 2.0° LR grid. The HR RCM grid is then trimmed so that both spatial dimensions are divisible by the downscaling factor of 4, producing paired LR-HR samples with grid sizes of 13 × 29 and 52 × 116, respectively. The model experiments use the period 1951--2099. The historical simulation is used for 1951--2005, and the RCP8.5 high-emissions scenario \cite{riahi2011rcp} is used for 2006--2099. This scenario is selected to increase the distributional shift between the historical training period and the future evaluation period. All GCM and RCM datasets choose the "r1i1p1" ensemble member.

Auxiliary geographic information includes a land-sea mask and elevation. The land-sea mask is derived from valid land pixels, while elevation is taken from the Parameter-elevation Regressions on Independent Slopes Model (PRISM) dataset \cite{daly2008physiographically} and interpolated from its original 4 km grid to the HR RCM grid. Ocean or invalid pixels are filled with zero and consistently masked across the LR and HR fields.

\section{Problem Formulation}

Let $r$ denote the spatial downscaling factor used in this study. For day $k$, let $\mathbf{x}_k \in \mathbb{R}^{1 \times h \times w}$ denote the LR GCM field and $\mathbf{y}_k \in \mathbb{R}^{1 \times H \times W}$ denote the corresponding HR RCM field, where $H=rh$ and $W=rw$. The downscaling task is to learn a parameterized mapping
\[
\hat{\mathbf{y}}_k = f_{\theta}(\mathbf{x}_k),
\]
where $\hat{\mathbf{y}}_k$ is the predicted HR field, $f_{\theta}$ is the downscaling model, and $\theta$ denotes trainable model parameters.

For the historical period, the HR field is used as the ground truth and to compute HR reconstruction loss; therefore, it is treated as the labeled source domain,
\[
\mathcal{D}_S = \{(\mathbf{x}_i^S, \mathbf{y}_i^S)\}_{i=1}^{N_S},
\]
where $S$ denotes source-domain samples, $i$ indexes historical training days, and $N_S$ is the number of labeled source samples. The future period is treated as the target domain,
\[
\mathcal{D}_T = \{\mathbf{x}_j^T\}_{j=1}^{N_T},
\]
where $T$ denotes target-domain samples, $j$ indexes future days, and $N_T$ is the number of target samples. During domain-adaptive training, the target HR fields $\mathbf{y}_j^T$ are not used. For ordinary supervised downscaling, the model is trained only on $\mathcal{D}_S$ by minimizing
\[
\mathcal{L}_{\mathrm{sup}}(\theta)
=
\frac{1}{N_S}
\sum_{i=1}^{N_S}
\left\|
f_{\theta}(\mathbf{x}_i^S) - \mathbf{y}_i^S
\right\|_2^2 ,
\]
where $\mathcal{L}_{\mathrm{sup}}$ is the supervised reconstruction loss and $\|\cdot\|_2^2$ denotes the sum of squared errors over the HR grid.

In a changing climate, the future target domain may not follow the same distribution as the historical source domain:
\[
P_S(\mathbf{x}, \mathbf{y}) \neq P_T(\mathbf{x}, \mathbf{y}),
\]
where $P_S$ and $P_T$ denote the joint source and target distributions of LR predictors and HR targets. This temporal distribution shift creates an OOD shift problem \cite{pan2009survey,ben2010theory}: a model trained only on historical LR-HR pairs may not maintain the same performance under future climate conditions.

To reduce this shift without using future HR labels during training, domain-adaptive downscaling combines the supervised source-domain reconstruction loss with an additional source-target alignment loss:
\[
\mathcal{L}_{\mathrm{DA}}(\theta)
=
\mathcal{L}_{\mathrm{sup}}(\theta)
+
\alpha \mathcal{L}_{\mathrm{align}}(\theta),
\]
where $\mathcal{L}_{\mathrm{align}}$ encourages the model to learn representations that are less sensitive to the historical-future distribution shift, and $\alpha$ controls the relative strength of this alignment term. The specific adversarial implementation of $\mathcal{L}_{\mathrm{align}}$ is described in the model architecture section.

\section{Model Architecture}

\begin{figure}[htbp]
    \centering
    \includegraphics[scale=0.4]{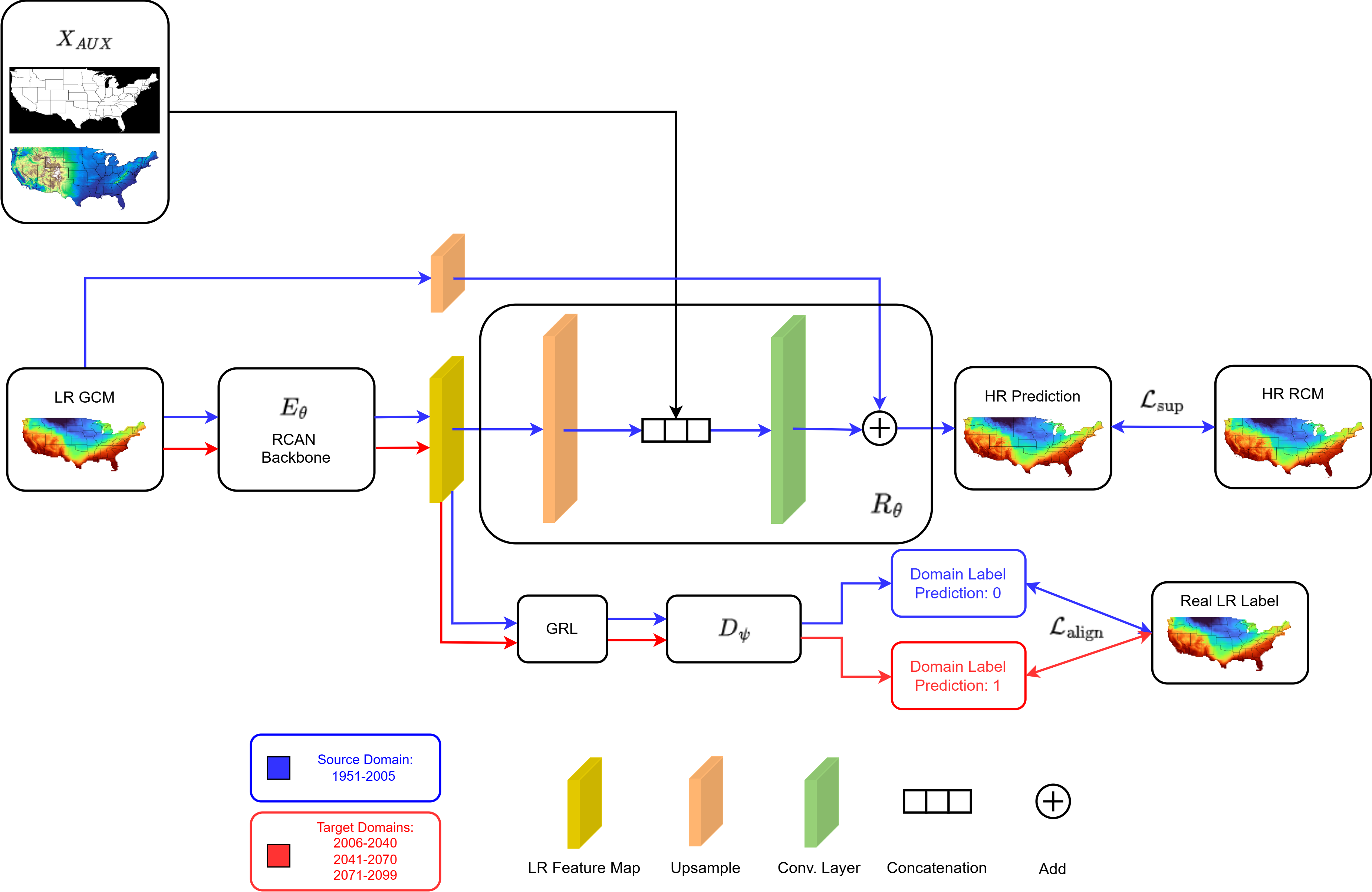}
\caption{Temporal domain-adaptive model for climate downscaling. The LR GCM input \(X_{\mathrm{LR}}\) is encoded by the RCAN feature extractor \(E_\theta\). For the supervised source-domain path, the extracted LR features are upsampled, concatenated with HR auxiliary fields \(X_{\mathrm{AUX}}\), and passed through the reconstruction head \(R_\theta\) to predict the HR RCM field. A bilinear skip connection from the LR input is added to the residual prediction. The supervised loss $\mathcal{L}_{\mathrm{sup}}$ is computed between the HR prediction and the HR RCM target. For domain adaptation, source and target LR feature maps are passed through a gradient reversal layer and domain classifier \(D_\psi\), producing source/target domain predictions trained with the adversarial domain loss $\mathcal{L}_{\mathrm{align}}$, where source samples are labeled 0 and target samples are labeled 1.}
\label{Fig:downscaling}
\end{figure}

The domain-adaptive downscaling model implements $f_{\theta}$ with two main components: an LR feature encoder $E_{\theta}$ and an HR residual reconstruction head $R_{\theta}$. The reconstruction head includes the learnable upsampling operation and the HR auxiliary-variable fusion. The domain classifier $D_{\psi}$ is attached to the encoder during training only and is removed during inference.

Given an LR GCM field $\mathbf{x} \in \mathbb{R}^{1 \times h \times w}$, the encoder first produces an LR feature map
\[
\mathbf{h} = E_{\theta}(\mathbf{x}).
\]
In this study, $E_{\theta}$ is based on residual channel-attention blocks (RCAN), which learn spatial patterns in the LR climate field while adaptively weighting informative feature channels \cite{zhang2018image}. As shown in Table \ref{tab:backbones}, we also tested alternative super-resolution backbones, including SRResNet \cite{ledig2017photo}, EDSR \cite{lim2017enhanced}, RRDB \cite{zhang2018residual}, and UNet \cite{ronneberger2015u}, but these architectures did not show a substantial performance advantage over the RCAN-based backbone.

Inside the HR residual reconstruction head, the encoded feature map is first upsampled to the HR grid:
\[
\mathbf{q} = U_{\theta}^{R}(\mathbf{h}),
\]
where $U_{\theta}^{R}$ denotes the pixel-shuffle upsampling block within $R_{\theta}$. We use pixel-shuffle upsampling because it learns feature-specific upscaling filters efficiently in LR feature space, unlike fixed bilinear interpolation, and avoids the uneven-overlap checkerboard artifacts often associated with transposed convolution \cite{shi2016real,odena2016deconvolution}. The upsampled feature map is concatenated with the HR auxiliary variables,
\[
\mathbf{z} = [\mathbf{q}, \mathbf{X_{AUX}}],
\]
where $[\cdot,\cdot]$ denotes channel-wise concatenation. This fusion allows the model to use land-sea and elevation information when reconstructing fine-scale temperature structure.

The complete reconstruction head then maps the encoded feature and auxiliary variables to a residual HR correction,
\[
\mathbf{r} = R_{\theta}(\mathbf{h},\mathbf{X_{AUX}}),
\]
where $R_{\theta}$ includes the internal upsampling step and the auxiliary fusion step. This residual correction is added to the bilinearly interpolated LR input:
\[
\hat{\mathbf{y}}
=
\mathcal{I}_{\mathrm{bilinear}}(\mathbf{x})
+
\mathbf{r}.
\]
Equivalently, the full downscaling pathway is
\[
\hat{\mathbf{y}}
=
\mathcal{I}_{\mathrm{bilinear}}(\mathbf{x})
+
R_{\theta}\left(E_{\theta}(\mathbf{x}),\mathbf{X_{AUX}}\right).
\]
This bilinear-plus-residual design is used because the interpolated LR GCM field already preserves the large-scale structure, while the neural network can focus on learning unresolved fine-scale corrections related to topography, coastlines, land-sea contrast, and local biases. It also reduces the need for the network to relearn the coarse-scale identity mapping and helps maintain consistency with the driving GCM signal. Similar interpolation-plus-correction designs have been widely used in image super-resolution \cite{kim2016accurate,kim2016deeply} and climate downscaling \cite{serifi2021spatio,park2026super}.

During domain-adaptive training, source and target samples share the same encoder:
\[
\mathbf{h}_i^S = E_{\theta}(\mathbf{x}_i^S),
\qquad
\mathbf{h}_j^T = E_{\theta}(\mathbf{x}_j^T).
\]
Only source samples are used for supervised reconstruction, while both source and target features are passed to the domain classifier. In this study, $D_{\psi}$ is implemented as a lightweight classifier that first applies average pooling over the encoded feature map and then passes the resulting feature vector through a multilayer perceptron with one hidden layer, ReLU activation, dropout, and a final two-unit output layer for source/target classification. This design can be written as
\[
D_{\psi}(\mathbf{h})
=
\mathrm{MLP}_{\psi}\left(\mathrm{AvgPool}(\mathbf{h})\right),
\]
where $\mathrm{AvgPool}(\cdot)$ denotes channel-wise averaging over all spatial locations in the encoded feature map. We also tested more complex domain classifiers with additional layers and larger hidden dimensions, but these alternatives did not provide further improvement, suggesting that the main role of $D_{\psi}$ is to provide a stable domain-alignment signal rather than to maximize classifier capacity.

The domain-classification dataflow is therefore
\[
\hat{\mathbf{d}}_i^S
=
D_{\psi}(\mathrm{GRL}_{\lambda}(\mathbf{h}_i^S)),
\qquad
\hat{\mathbf{d}}_j^T
=
D_{\psi}(\mathrm{GRL}_{\lambda}(\mathbf{h}_j^T)).
\]
Here, $\hat{\mathbf{d}}$ is the predicted domain label. The adversarial alignment loss is implemented as
\[
\mathcal{L}_{\mathrm{align}}(\theta,\psi)
=
\frac{1}{N_S}
\sum_{i=1}^{N_S}
\mathrm{CE}\left(D_{\psi}(\mathrm{GRL}_{\lambda}(\mathbf{h}_i^S)),0\right)
+
\frac{1}{N_T}
\sum_{j=1}^{N_T}
\mathrm{CE}\left(D_{\psi}(\mathrm{GRL}_{\lambda}(\mathbf{h}_j^T)),1\right),
\]
where $\mathrm{CE}(\cdot,\cdot)$ is the cross-entropy loss, labels 0 and 1 indicate the historical source and future target domains, respectively, and $\mathrm{GRL}_{\lambda}$ is the gradient reversal layer with coefficient $\lambda$. The classifier learns to distinguish historical from future features, while the gradient reversal layer leaves features unchanged in the forward pass but multiplies the encoder gradient by $-\lambda$ during backpropagation, encouraging the encoder to make source and target features more domain-invariant \cite{ganin2016domain}. At inference time, $D_{\psi}$ and $\mathrm{GRL}_{\lambda}$ are discarded, and a future LR GCM field is downscaled as
\[
\hat{\mathbf{y}}_j^T
=
\mathcal{I}_{\mathrm{bilinear}}(\mathbf{x}_j^T)
+
R_{\theta}\left(E_{\theta}(\mathbf{x}_j^T),\mathbf{X_{AUX}}\right).
\]

\section{Baseline Models}

To evaluate the proposed domain-adaptive model, we compare it with interpolation, statistical bias-correction, and DL baselines. These methods represent different levels of model complexity, from non-learning interpolation to distributional correction and deep super-resolution.

\textbf{Bilinear Interpolation}. The simplest baseline directly interpolates the LR GCM field to the HR grid:
\[
\hat{\mathbf{y}}=\mathcal{I}_{\mathrm{bilinear}}(\mathbf{x}).
\]
This method provides a reference for how much improvement is gained beyond smooth spatial interpolation, which is commonly used as a simple baseline in super-resolution and climate downscaling studies \cite{vandal2017deepsd,bano2020configuration}. It preserves the large-scale GCM pattern but cannot recover unresolved terrain-driven error, coastal fine-scale variability, or other inherent biases between climate models.

\textbf{BCSD}. Bias Correction and Spatial Disaggregation (BCSD) first interpolates the LR GCM field to the HR grid and then applies a monthly, grid-cell-wise correction based on historical model-target statistics \cite{wood2004hydrologic}. Let $z=\mathcal{I}_{\mathrm{bilinear}}(\mathbf{x})$ denote the interpolated GCM value for a given month $m$ and HR grid cell. The corrected value is
\[
\hat{y}
=
\frac{z-\mu_{z,m}}{\sigma_{z,m}}
\sigma_{y,m}
+
\mu_{y,m},
\]
where $\mu_{z,m}$ and $\sigma_{z,m}$ are the historical mean and standard deviation of the interpolated GCM field, and $\mu_{y,m}$ and $\sigma_{y,m}$ are the corresponding statistics of the HR RCM target.

\textbf{QDM}. Quantile Delta Mapping (QDM) is a distributional bias-correction method designed to correct historical biases while preserving the modeled climate-change signal \cite{cannon2015bias}. For a future interpolated GCM value $z$, QDM estimates its quantile level $p$ and applies an additive correction:
\[
\hat{y}
=
F^{-1}_{y_S,m}(p)
+
\left[
z-F^{-1}_{z_S,m}(p)
\right],
\]
where $F^{-1}_{y_S,m}$ and $F^{-1}_{z_S,m}$ are the historical target and historical model quantile functions for month $m$. This makes QDM more suitable than simple quantile mapping when preserving projected changes is important.

\textbf{CDF-t}. CDF-t is another empirical distribution-transfer method that uses cumulative distribution functions to relate historical model, historical target, and future model distributions \cite{michelangeli2009probabilistic}. In this study, CDF-t is applied month-wise and grid-cell-wise to transfer the historical model-target CDF relationship to future GCM inputs. Compared with mean-variance correction, it can correct more of the distributional shape, although it still relies on empirical historical relationships.

\textbf{RCAN}. The DL baseline uses the same encoder-reconstruction pathway described above but removes the domain classifier and the alignment loss. It is trained only with the supervised reconstruction loss on historical LR-HR pairs. This baseline tests whether a supervised deep downscaling model can learn nonlinear fine-scale corrections without explicit temporal domain adaptation.

\textbf{RCAN-QDM}. This baseline applies QDM as a post-processing step to the RCAN predictions. The correction is fitted using historical RCAN outputs and historical HR RCM targets, and is then applied to future RCAN predictions. This method combines nonlinear spatial reconstruction from RCAN with distributional adjustment from QDM \cite{cannon2015bias}, and was used in previous downscaling research as a trend-preserving technique \cite{wang2024multivariate,lin2023deep}.

\textbf{RCAN-GAN}. This baseline uses the RCAN model as the generator and adds a conditional PatchGAN \cite{demir2018patch} discriminator during training. The discriminator is conditioned on the candidate HR field, the bilinearly upsampled LR GCM input, and the HR auxiliary channels. It is implemented as a patch-level convolutional classifier. Therefore, instead of producing a single image-level real/fake score, it outputs a grid of patch logits that judge whether local HR structures are realistic for the corresponding coarse-scale forcing and geographic context. Training follows the general GAN framework \cite{goodfellow2014generative} and conditional image-to-image or super-resolution designs \cite{ledig2017photo,isola2017image}. In the implementation, the generator is first warmed up with supervised losses, while the discriminator is trained with the corresponding least-squares GAN objective. Recent GAN-based downscaling and climate post-processing studies suggest that these models can better preserve spatial structure and climate-change signals under future or non-stationary conditions \cite{abdelmoaty2026does,pan2021learning,rampal2025reliable}.

\section{Training and Evaluation Metrics}
We define the 1951--2005 historical period as the training set and divide the RCP8.5 simulation into three validation sets: 2006--2040, 2041--2070, 2071--2099, representing the current, mid-century, and end-of-century climate. The historical period is used as the labeled source domain, while the three future periods are used as the target domains. Before training, the LR input and HR target fields are normalized by subtracting the mean and dividing by the standard deviation using statistics computed only from the source training period. The same source-domain mean and standard deviation are then applied to the target-period inputs and validation targets. 

For each training iteration, one labeled source mini-batch and one unlabeled target mini-batch are sampled. The source batch is used for supervised downscaling, while both source and target batches are used for adversarial domain alignment. The supervised reconstruction loss is the mean squared error between the normalized prediction and normalized HR RCM target. The domain classifier is trained with cross-entropy loss to distinguish source from target features. The gradient reversal coefficient is increased during training according to
\[
\lambda(p)
=
\lambda_{\max}
\left(
\frac{2}{1+\exp(-10p)}-1
\right),
\]
where $p$ is the normalized training progress and $\lambda_{\max}=1$.

Model selection is performed first for the supervised RCAN model. Model parameters are optimized with AdamW, with gradient norms clipped to 1.0 and automatic mixed precision used when enabled on GPU. The learning rate is adjusted using ReduceLROnPlateau with a patience of 10 epochs and a decay factor of 0.5. To prevent future HR targets from influencing model selection, Optuna \cite{akiba2019optuna} is applied using a chronological 80\%--20\% split of the historical period: 1951--1994 is used for supervised training, and 1995--2005 is reserved for historical validation. Optuna selects the RCAN architecture and shared training hyperparameters by minimizing the normalized MSE over the historical validation subset. The selected configuration is then frozen, and the final RCAN model is retrained using all historical LR--HR pairs from 1951--2005 before evaluation on the future periods.

RCAN-DA uses the RCAN architecture and shared training hyperparameters selected above. Domain-adaptation-specific hyperparameters, including the domain-loss weight and domain-classifier configuration, are selected using the same historical split. The LR--HR pairs in the training subset form the labeled source domain, while the LR fields in the validation subset are treated as unlabeled pseudo-target samples for domain alignment. Their corresponding HR fields are used only to calculate the Optuna validation objective and do not contribute to gradient-based training. After the domain-adaptation-specific hyperparameters are frozen, the final RCAN-DA model is trained using the complete historical record as the labeled source domain and LR fields from the corresponding future period as the unlabeled target domain. Future HR fields are not used for model training, hyperparameter selection, learning-rate scheduling, or early stopping and are accessed only for final evaluation. During final training, the learning-rate scheduler monitors the source-domain reconstruction loss. A separate RCAN-DA model is trained for each future target period. The complete hyperparameter search space is shown in Table \ref{tab:hyperparameter_space}.

\begin{table}[t]
\centering
\caption{Hyperparameter search space and selected values of the proposed model.}
\label{tab:hyperparameter_space}
\begin{tabular}{lll}
\hline
\textbf{Hyperparameter} & \textbf{Search space} & \textbf{Best} \\
\hline
Batch size & $\{128, 256, 512\}$ & 512 \\
Learning rate & $[10^{-5}, 3 \times 10^{-4}]$ log-uniform & $1 \times 10^{-4}$ \\
Weight decay & $[10^{-7}, 10^{-3}]$ log-uniform & $8 \times 10^{-5}$ \\
Number of residual blocks & $\{4, 6, 8, 10, 12, 14, 16\}$ & 12 \\
Number of feature channels & $\{64, 96, 128\}$ & 128 \\
Number of RCAN groups & $\{1, 2, 4, 8\}$ & 8 \\
Channel-attention reduction ratio & $\{4, 8, 16\}$ & 16 \\
Residual scaling factor & $\{0.2, 0.3, 0.4\}$ & 0.3 \\
Domain-loss weight $\alpha$ & $[10^{-5}, 10^{-1}]$ log-uniform & $5 \times 10^{-4}$ \\
Domain-classifier hidden dimension & $\{64, 128, 256\}$ & 64 \\
\hline
\end{tabular}
\end{table}

Model performance is evaluated using mean squared error (MSE), root mean squared error (RMSE), peak signal-to-noise ratio (PSNR), and structural similarity index measure (SSIM). PSNR and SSIM are commonly used in image super-resolution studies to evaluate pixel-level reconstruction accuracy and spatial structural similarity. Let $\mathcal{S}$ denote the evaluation sample set, with $|\mathcal{S}|=M$, and let $\Omega$ denote the HR spatial grid, with $|\Omega|=P$. For sample $i \in \mathcal{S}$ and grid cell $p \in \Omega$, $\hat{y}_{i,p}$ and $y_{i,p}$ denote the predicted and target HR temperature values, respectively.

The MSE is computed as
\begin{equation}
\mathrm{MSE} =
\frac{1}{MP}
\sum_{i \in \mathcal{S}}
\sum_{p \in \Omega}
\left(\hat{y}_{i,p} - y_{i,p}\right)^2 .
\end{equation}

RMSE is the square root of MSE and is reported in the same physical unit as temperature:
\begin{equation}
\mathrm{RMSE} =
\sqrt{\mathrm{MSE}} .
\end{equation}

PSNR measures reconstruction quality relative to the dynamic range of the target HR temperature field:
\begin{equation}
\mathrm{PSNR} =
10 \log_{10}
\left(
\frac{L^2}{\mathrm{MSE}}
\right),
\end{equation}
where $L = y_{\max} - y_{\min}$ is the target temperature range in the evaluation period.

SSIM measures the structural similarity between the predicted and target HR fields \cite{wang2004image}. For a local window, SSIM is defined as
\begin{equation}
\mathrm{SSIM}(\hat{\mathbf{y}}, \mathbf{y}) =
\frac{
(2\mu_{\hat{y}}\mu_y + C_1)(2\sigma_{\hat{y}y} + C_2)
}{
(\mu_{\hat{y}}^2 + \mu_y^2 + C_1)(\sigma_{\hat{y}}^2 + \sigma_y^2 + C_2)
},
\end{equation}
where $\mu_{\hat{y}}$ and $\mu_y$ are local means, $\sigma_{\hat{y}}^2$ and $\sigma_y^2$ are local variances, and $\sigma_{\hat{y}y}$ is the local covariance between prediction and target. The constants are defined as $C_1=(0.01L)^2$ and $C_2=(0.03L)^2$. In this study, SSIM is computed using an $11 \times 11$ moving window and averaged over all evaluation samples and grid cells. Lower MSE and RMSE and higher PSNR and SSIM indicate better downscaling performance.

\section{Result}
\subsection{Temporal Distribution Shift in Future Climate Conditions}

\begin{figure}[htbp]
    \centering
    \includegraphics[scale=0.6]{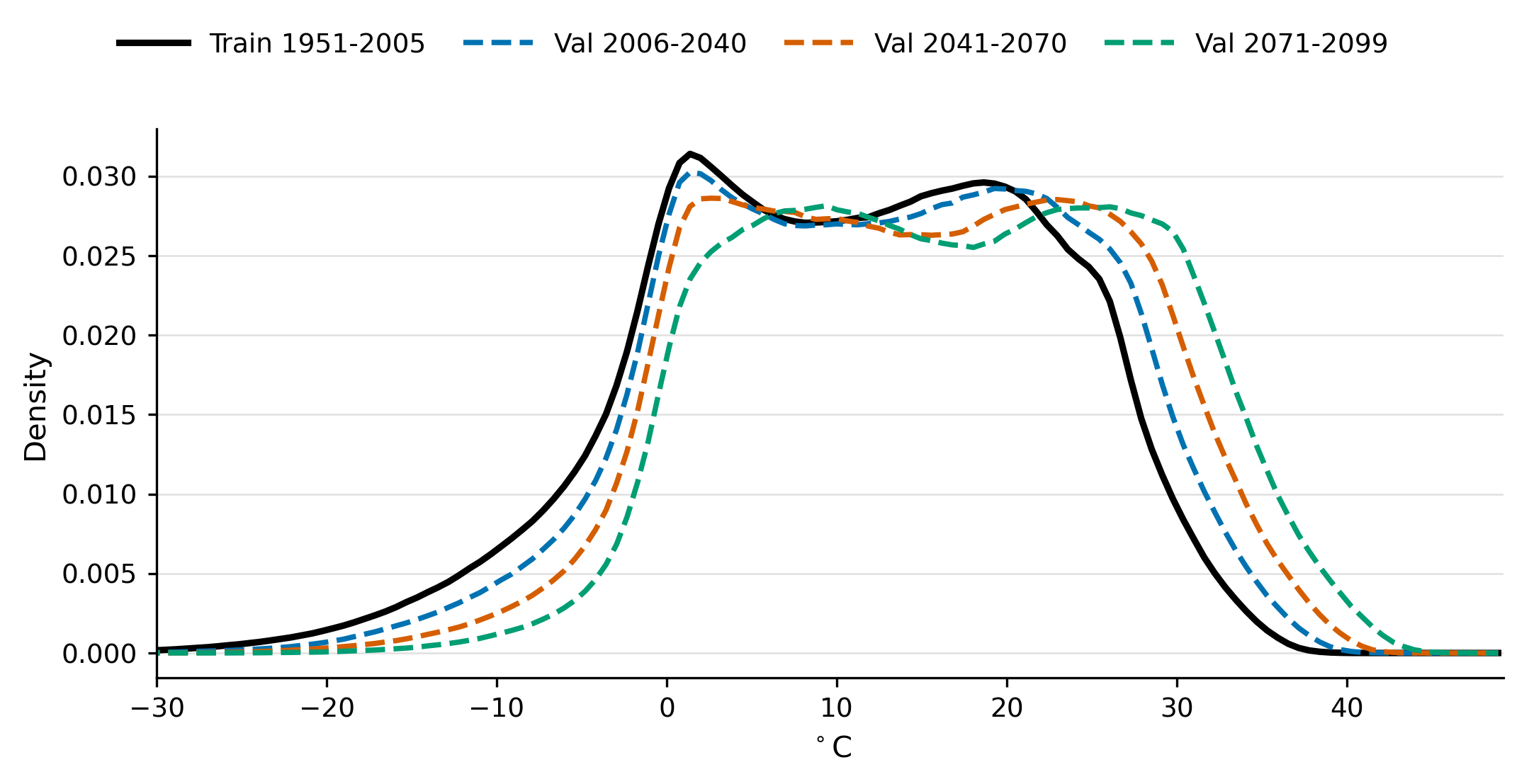}
\caption{Marginal distribution shift of daily near-surface air temperature over CONUS for the HR RCM target field. The dataset for this figure is CanESM2-RCA4. The historical training period covers 1951--2005, while the validation periods are split into 2006--2040, 2041--2070, and 2071--2099. The progressive rightward shift of the distributions indicates warming and increasing temporal OOD differences between the training and future validation periods.}
\label{Fig:train_val_temperature_distribution}
\end{figure}

Before evaluating the downscaling models, we first examine whether the validation periods represent a temporally shifted climate relative to the historical training period. Figure \ref{Fig:train_val_temperature_distribution} shows the marginal distributions of daily near-surface air temperature over CONUS for the training period and three future validation periods. Compared with the 1951--2005 training distribution, the validation distributions progressively shift toward higher temperatures, with the largest displacement occurring in 2071--2099. This indicates that the model evaluation is not only testing interpolation within the historical climate regime, but also testing adaptation under increasingly non-stationary future conditions. Table \ref{tab:temperature_distribution_summary} summarizes the daily temperature statistics of the HR RCM target fields used in this study. The table reports results for three RCM simulations, identified by their driving GCM and RCM combination. For each RCM target dataset, future periods show positive mean temperature shifts relative to the historical training period, while changes in the median and standard deviation indicate that the shift affects the broader target distribution rather than only isolated extremes. These results confirm that the validation periods introduce a clear temporal OOD challenge for supervised downscaling models trained on historical LR-HR pairs.

\begin{figure}[htbp]
    \centering
    \includegraphics[scale=0.51]{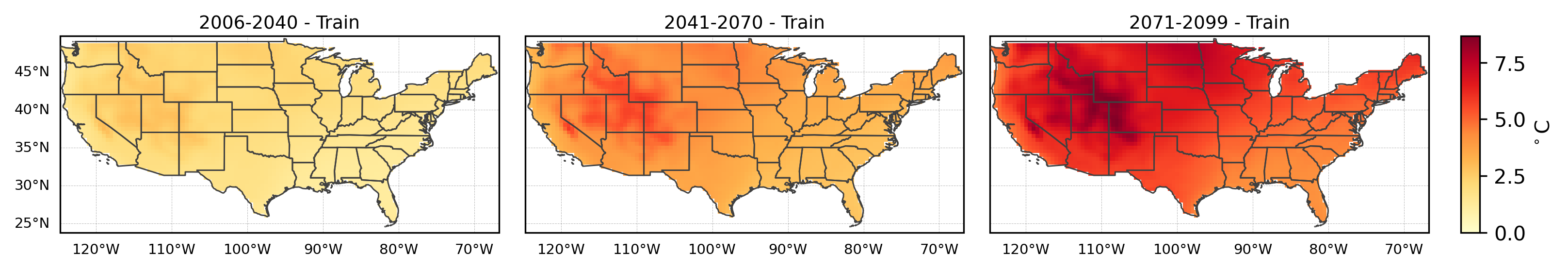}
\caption{Spatial pattern of mean temperature change over CONUS relative to the historical training period. The dataset for this figure is CanESM2-RCA4. Each panel shows the difference between the mean HR RCM temperature in a future validation period and the mean temperature during the training period.}
\label{Fig:future_minus_historical_temperature_maps}
\end{figure}

The spatial structure of this warming is shown in Figure \ref{Fig:future_minus_historical_temperature_maps}. All future periods show positive mean temperature differences relative to the training period, but the magnitude and spatial extent increase over time. The strongest warming appears in the later validation periods, especially across parts of the western and central CONUS. This spatially heterogeneous warming pattern further motivates evaluating whether downscaling models can preserve regional climate structure while adapting beyond the historical training distribution.

\subsection{Downscaling Performance Under Future Climate Conditions}

\begin{table}[htbp]
\centering
\caption{Validation-period downscaling performance for daily near-surface air temperature over CONUS using the CanESM2-RCA4 simulation. Metrics are computed against the HR RCM target fields for three future validation periods. Bold values indicate the best-performing method for each metric within each validation period.}
\label{tab:canesm2_rca4_validation_metrics}
\begin{tabular}{llrrrr}
\hline
Period & Method & MSE & RMSE & PSNR & SSIM \\
\hline
\multirow{8}{*}{2006--2040}
& Bilinear & 31.946 & 5.652 & 23.527 & 0.60087 \\
& BCSD & 5.883 & 2.425 & 30.875 & 0.86848 \\
& QDM & 6.119 & 2.474 & 30.704 & 0.86416 \\
& CDF-t & 6.668 & 2.582 & 30.331 & 0.85684 \\
& RCAN & 4.552 & 2.134 & 31.989 & 0.88810 \\
& RCAN-QDM & 4.932 & 2.221 & 31.641 & 0.89023 \\
& RCAN-GAN & 4.494 & 2.120 & 32.045 & 0.88888 \\
& RCAN-DA & \textbf{4.145} & \textbf{2.036} & \textbf{32.395} & \textbf{0.89824} \\
\hline
\multirow{8}{*}{2041--2070}
& Bilinear & 34.391 & 5.864 & 22.944 & 0.60852 \\
& BCSD & 5.968 & 2.443 & 30.550 & 0.87378 \\
& QDM & 6.674 & 2.583 & 30.065 & 0.85898 \\
& CDF-t & 7.010 & 2.648 & 29.851 & 0.85882 \\
& RCAN & 4.643 & 2.155 & 31.641 & 0.89488 \\
& RCAN-QDM & 4.933 & 2.221 & 31.377 & 0.89619 \\
& RCAN-GAN & 4.554 & 2.134 & 31.724 & 0.89611 \\
& RCAN-DA & \textbf{4.174} & \textbf{2.043} & \textbf{32.103} & \textbf{0.90358} \\
\hline
\multirow{8}{*}{2071--2099}
& Bilinear & 38.200 & 6.181 & 22.170 & 0.61061 \\
& BCSD & 6.300 & 2.510 & 29.998 & 0.87975 \\
& QDM & 8.123 & 2.850 & 28.894 & 0.84555 \\
& CDF-t & 7.541 & 2.746 & 29.217 & 0.86137 \\
& RCAN & 5.008 & 2.238 & 30.994 & 0.90157 \\
& RCAN-QDM & 5.174 & 2.275 & 30.853 & 0.90133 \\
& RCAN-GAN & 4.908 & 2.215 & 31.082 & 0.90187 \\
& RCAN-DA & \textbf{4.387} & \textbf{2.094} & \textbf{31.570} & \textbf{0.90682} \\
\hline
\end{tabular}
\end{table}

We next evaluate the downscaling accuracy of all models across the three future validation periods. Table \ref{tab:canesm2_rca4_validation_metrics} reports the main quantitative results for the CanESM2-RCA4 simulation, while the corresponding results for the other two GCM-RCM pairs are provided in Tables \ref{tab:canesm2_canrcm4_validation_metrics} and \ref{tab:ecearth_rca4_validation_metrics}. Overall, validation performance tends to degrade toward the end of the century, indicating that downscaling becomes more difficult as the evaluation climate moves farther from the historical training distribution. Across all validation periods, bilinear interpolation produces the largest errors and the lowest structural similarity, showing that simple spatial interpolation cannot recover the fine-scale temperature structure of the HR RCM field. The large bilinear errors also reflect the discrepancy between the LR GCM predictor and the HR RCM target, including biases that can arise across GCM-RCM model chains \cite{sorland2018bias}. Statistical bias-correction baselines substantially reduce the error relative to bilinear interpolation, but they remain less accurate than the DL models. Applying QDM after the DL model (RCAN-QDM) does not improve performance in these experiments, and instead produces a slight degradation relative to the standard RCAN. The GAN-based model generally outperforms the baseline RCAN, suggesting that adversarial training can help under non-stationary climate conditions. The domain-adaptive model achieves the best score for every metric and validation period, with the lowest MSE and RMSE and the highest PSNR and SSIM. The magnitude of improvement also increases in the more strongly shifted periods. In the near-future period, 2006--2040, RCAN-DA reduces MSE by about 9\% relative to RCAN. In the end-of-century climate, 2071--2099, RCAN-DA reduces MSE from 5.008 to 4.387, corresponding to about a 12\% reduction. Similar patterns are observed for the other two datasets, suggesting that domain adaptation improves temporal generalization across different GCM-RCM pairs.

\begin{figure}[htbp]
    \centering
    \includegraphics[scale=0.33]{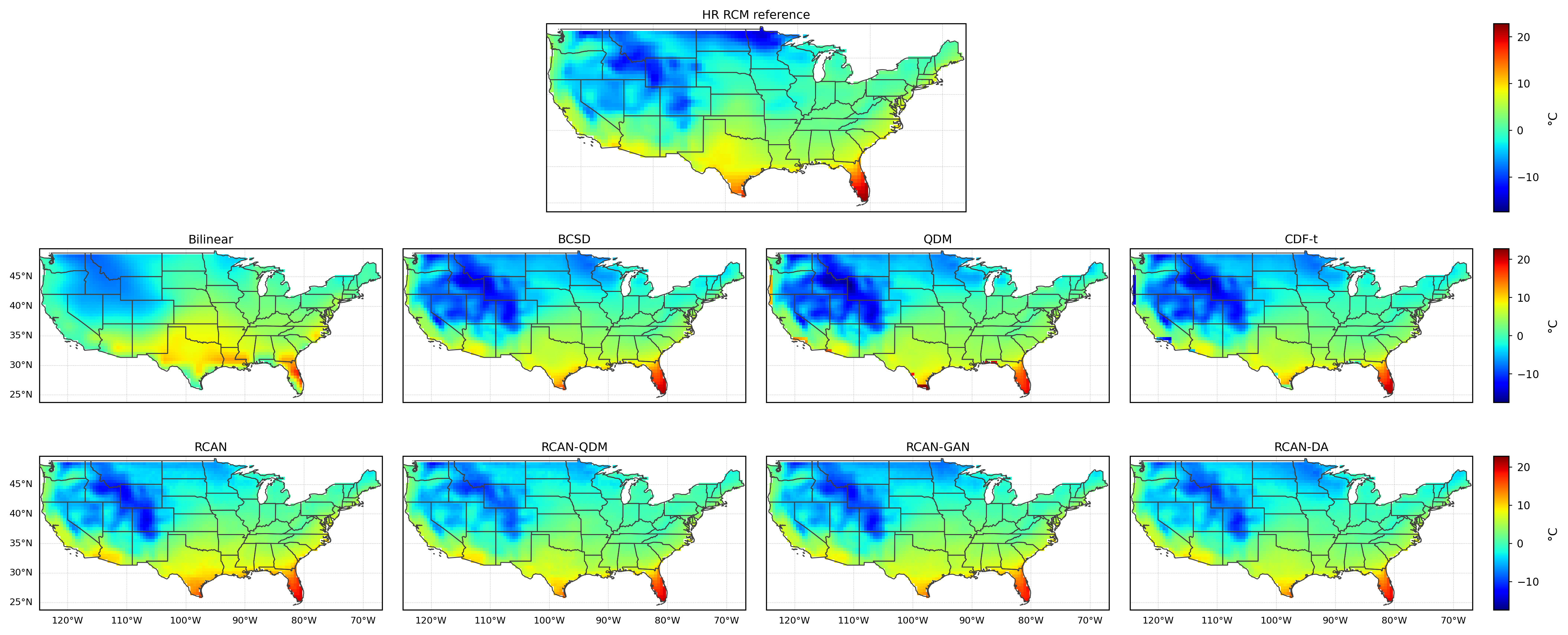}
\caption{Downscaled daily near-surface air temperature fields. The top panel shows the HR RCM reference, and the remaining panels show predictions from bilinear interpolation, statistical downscaling baselines, and DL-based models.}
\label{Fig:prediction_value_maps}
\end{figure}

Figure~\ref{Fig:prediction_value_maps} shows an example of the spatial prediction fields. Bilinear interpolation produces overly smooth fields and weakens regional gradients, especially along the western topographic regions, including the Pacific Coast, Rocky Mountains, and interior western United States. Other statistical methods recover more spatial variability than interpolation but still differ from the HR RCM reference in several regions, including the northern Rocky Mountains, northern Great Plains, and Upper Midwest. The DL models more closely reproduce the large-scale temperature pattern and fine-scale gradients of the HR RCM target, particularly the warm conditions over the southern Great Plains, Gulf Coast, and Florida. Among them, RCAN-DA provides the strongest quantitative performance while maintaining a visually consistent spatial structure, supporting the idea that domain adaptation helps preserve the LR-to-HR mapping under future climate conditions. However, not all RCM features are fully recovered. In particular, the strong cold feature over the northern Rocky Mountains, northern Great Plains, and parts of the Upper Midwest is difficult for the downscaling models to reproduce, likely because this feature is weak or absent in the corresponding GCM input field.

The absolute difficulty of the task varies across the three simulations. CanESM2-CanRCM4 shows the lowest errors overall, possibly because the driving GCM and RCM come from the same modeling center and may have more compatible model configurations. In contrast, EC-EARTH-RCA4 has substantially larger MSE and RMSE values, indicating a more challenging downscaling target. Despite this dataset dependence, the relative model ranking remains stable: domain adaptation consistently outperforms the standard DL models, bias-correction methods and the classical baselines across the future validation periods.

\begin{figure}[htbp]
    \centering
    \includegraphics[scale=0.4]{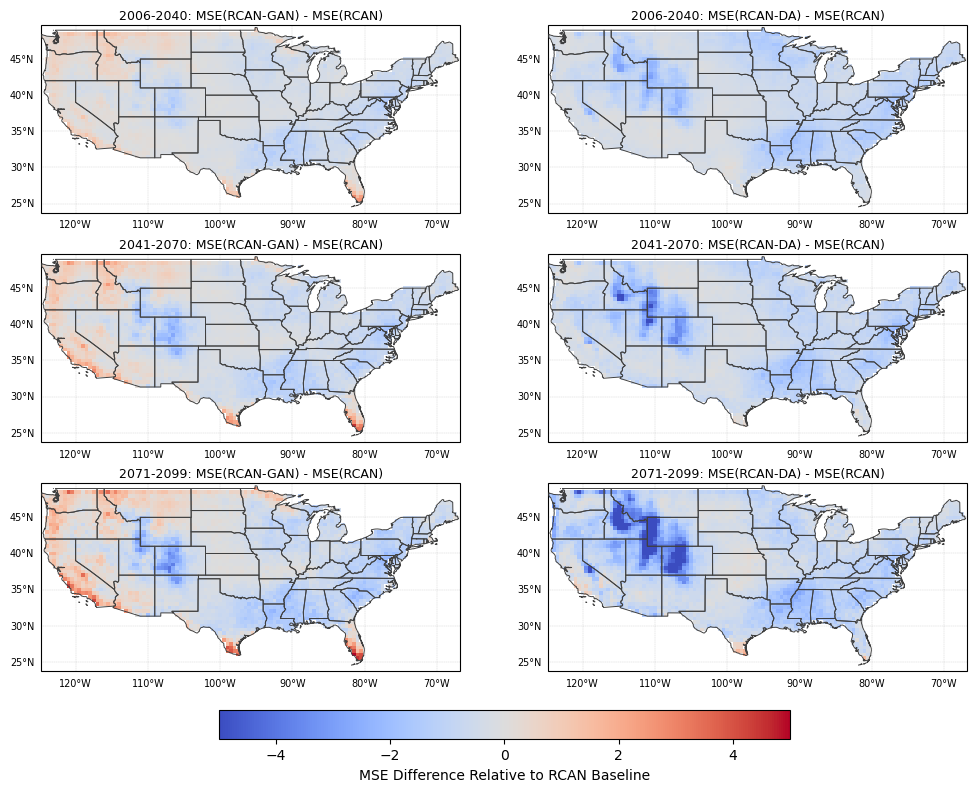}
\caption{Spatial differences in MSE relative to the standard RCAN model for the CanESM2-RCA4 dataset. Each panel shows the local MSE difference between either RCAN-GAN or RCAN-DA and RCAN for one validation period, computed as $\mathrm{MSE}_{\mathrm{model}}-\mathrm{MSE}_{\mathrm{RCAN}}$. Blue colors indicate regions where the model improves over RCAN, while red colors indicate larger errors than RCAN.}
\label{Fig:error_improvement_maps}
\end{figure}

To further examine where the domain-adaptive model improves over the standard RCAN, Figure \ref{Fig:error_improvement_maps} shows spatial differences in local MSE for RCAN-GAN and RCAN-DA relative to RCAN across the three validation periods. Negative values indicate regions where the modified model reduces error compared with standard RCAN, while positive values indicate increased error. RCAN-DA shows widespread error reductions across much of CONUS, with particularly strong improvements over the western mountainous areas. The magnitude and spatial extent of these improvements increase in the later validation periods, especially in 2071--2099, consistent with the quantitative results in Table \ref{tab:canesm2_rca4_validation_metrics}. On the other hand, RCAN-GAN also reduces errors in several regions, including parts of the central and eastern United States, but the magnitude of improvement does not change much when the warming signal is strong and it shows localized error increases along parts of the Pacific Coast, the Southwest, Texas, and Florida. This suggests that adversarial training can improve spatial realism in some areas but does not provide the same consistent temporal adaptation as the proposed approach.

\begin{figure}[htbp]
    \centering
    \includegraphics[scale=0.55]{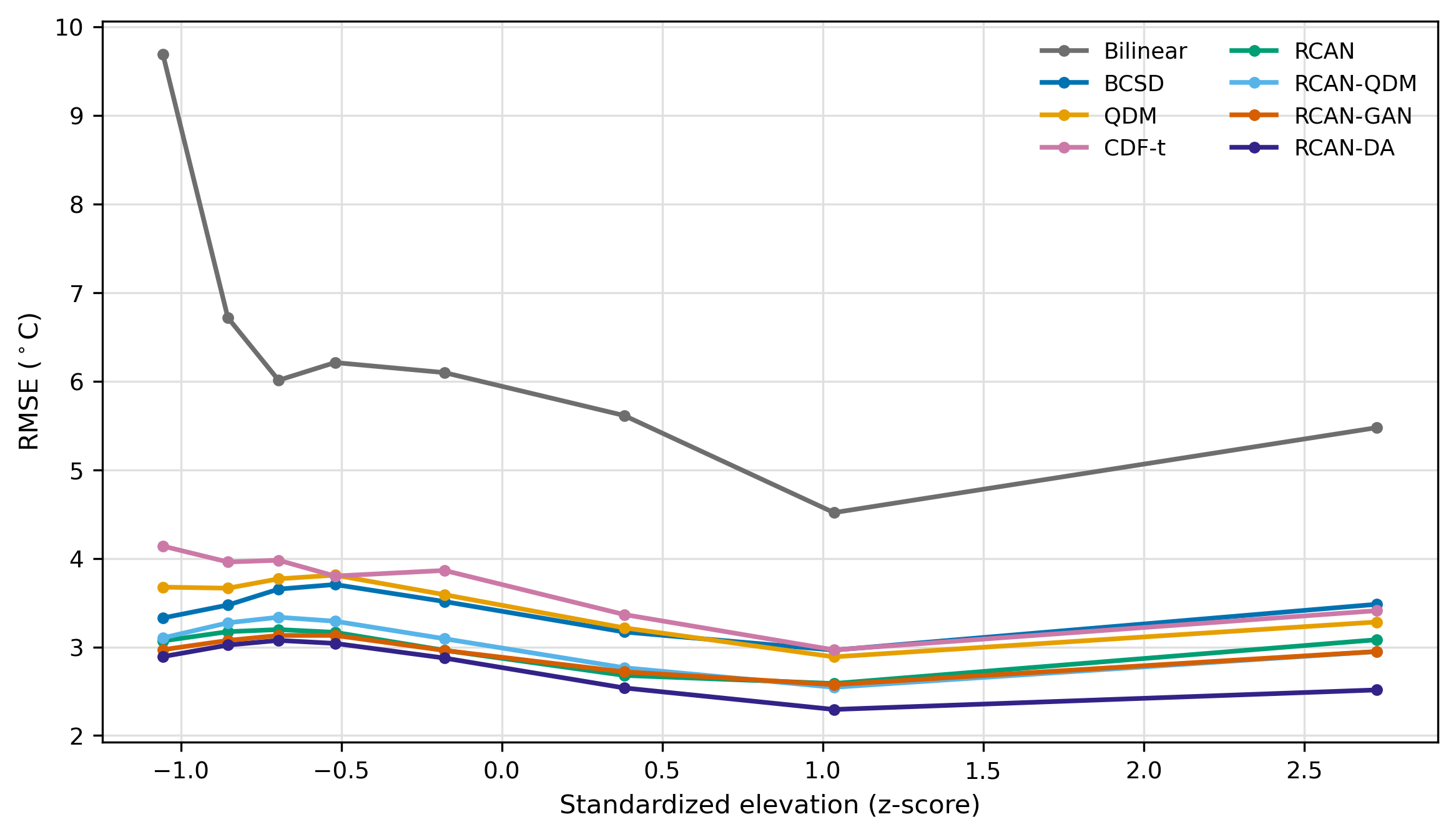}
\caption{RMSE of downscaled daily near-surface air temperature across standardized elevation bins for the CanESM2-RCA4 dataset during the 2071--2099 validation period. Elevation is standardized over CONUS land pixels using the z-score method, where zero corresponds to the mean of approximately 784\,m. Each bin contains the same number of sampled pixels.}
\label{Fig:fig_error_by_elevation}
\end{figure}

Figure \ref{Fig:fig_error_by_elevation} further examines model performance across different topographic regimes. Bilinear interpolation produces substantially larger errors across all elevation ranges, with particularly high RMSE over the lowest-elevation bin, likely reflecting difficulties near coastlines and regions with strong land-sea contrasts. The statistical and DL methods considerably reduce these errors, although most exhibit a U-shaped pattern, with lower RMSE at intermediate elevations and increasing error at the highest elevations, where topographic effects are more complex. Among the DL models, RCAN-DA achieves the lowest RMSE in every elevation bin. Its improvement over the standard RCAN becomes more obvious at higher elevations: in the highest-elevation bin, RCAN-DA reduces RMSE from approximately 3.1\,\textdegree C to 2.5\,\textdegree C. RCAN-GAN also improves slightly over RCAN at higher elevations, but the reduction is smaller. These results complement the spatial error-difference maps by showing that the strongest benefits of domain adaptation occur in elevated terrain, particularly across the Rocky Mountains and other mountainous regions of the western CONUS, where the coarse GCM input provides limited representation of local topographic temperature gradients.

\begin{figure}[htbp]
    \centering
    \includegraphics[scale=0.55]{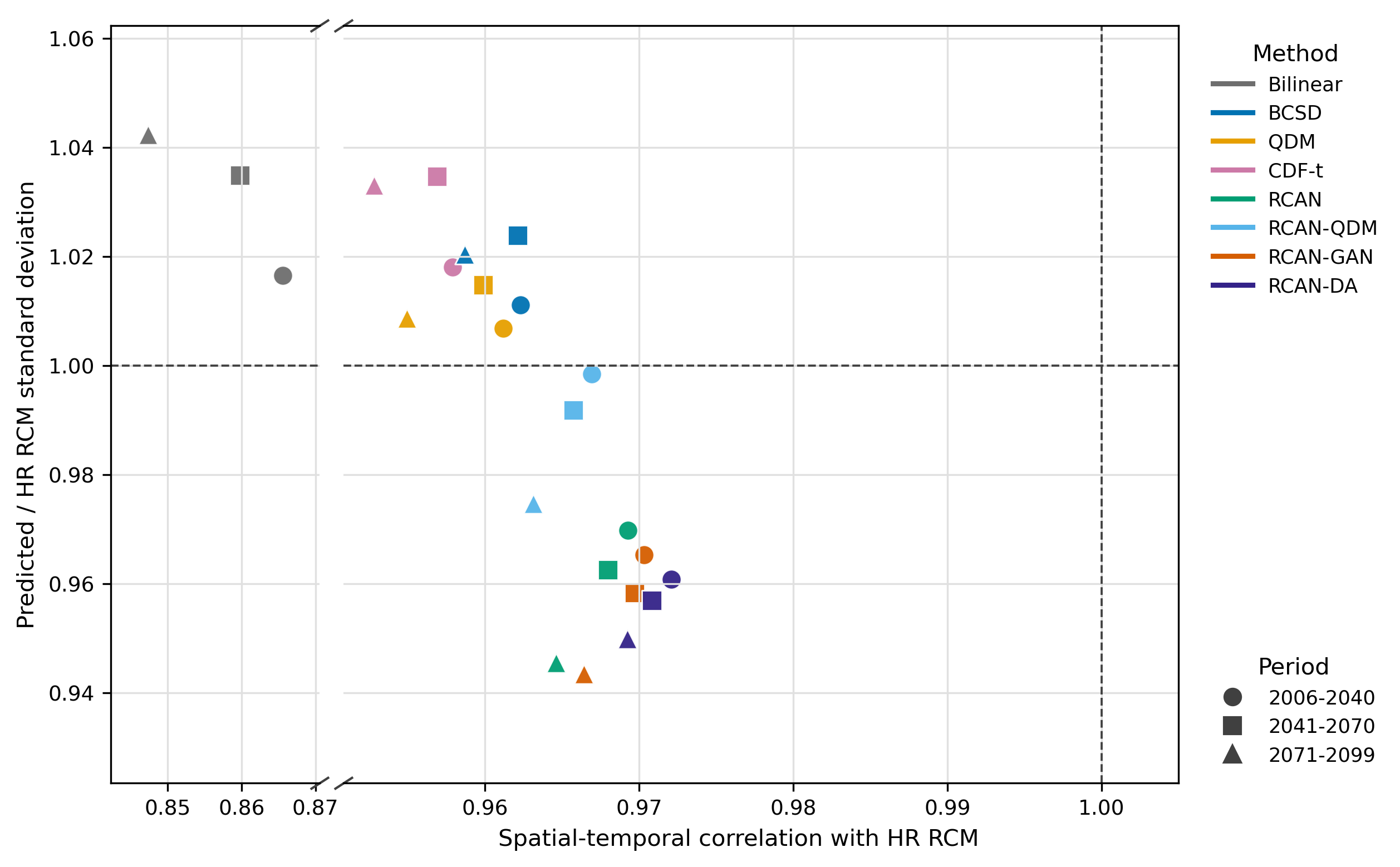}
\caption{Taylor-style diagnostics for daily near-surface air temperature over CONUS using the CanESM2-RCA4 dataset. The horizontal axis shows the spatial-temporal correlation between each downscaled field and the HR RCM target, while the vertical axis shows the ratio of predicted to RCM standard deviation and therefore measures how well each method reproduces the magnitude of target temperature variability. A ratio of one indicates correctly reproduced variability, values below one indicate underestimated variability, and values above one indicate excessive variability.}
\label{Fig:fig_taylor_style_diagnostics}
\end{figure}

We also compare how well each method reproduces the HR RCM's spatial-temporal pattern and variability across the three validation periods using a Taylor-style diagnostic plot. Figure \ref{Fig:fig_taylor_style_diagnostics} shows that bilinear interpolation has substantially lower correlation with the HR RCM target than all statistical and learning-based methods, despite producing a standard deviation slightly above the reference value. The statistical methods increase correlation considerably but generally overestimate variability, particularly CDF-t. The DL-based models achieve the highest correlations, with RCAN-DA consistently providing the strongest correlation across all three validation periods. However, RCAN, RCAN-GAN, and RCAN-DA have standard-deviation ratios below one, indicating that their predictions remain smoother and less variable than the HR RCM fields. RCAN-QDM most closely reproduces the target variance, although its correlation is slightly lower than that of RCAN-DA. For most DL-based models, both correlation and variance reproduction deteriorate modestly toward 2071--2099, reflecting the increasing difficulty of adaptation under stronger temporal distribution shift. Thus, RCAN-DA best preserves the spatial-temporal pattern of the HR target, while RCAN-QDM provides the closest match to its overall variability.

\subsection{Bias and Extremes Analysis}

\begin{figure}[htbp]
    \centering
    \includegraphics[scale=0.55]{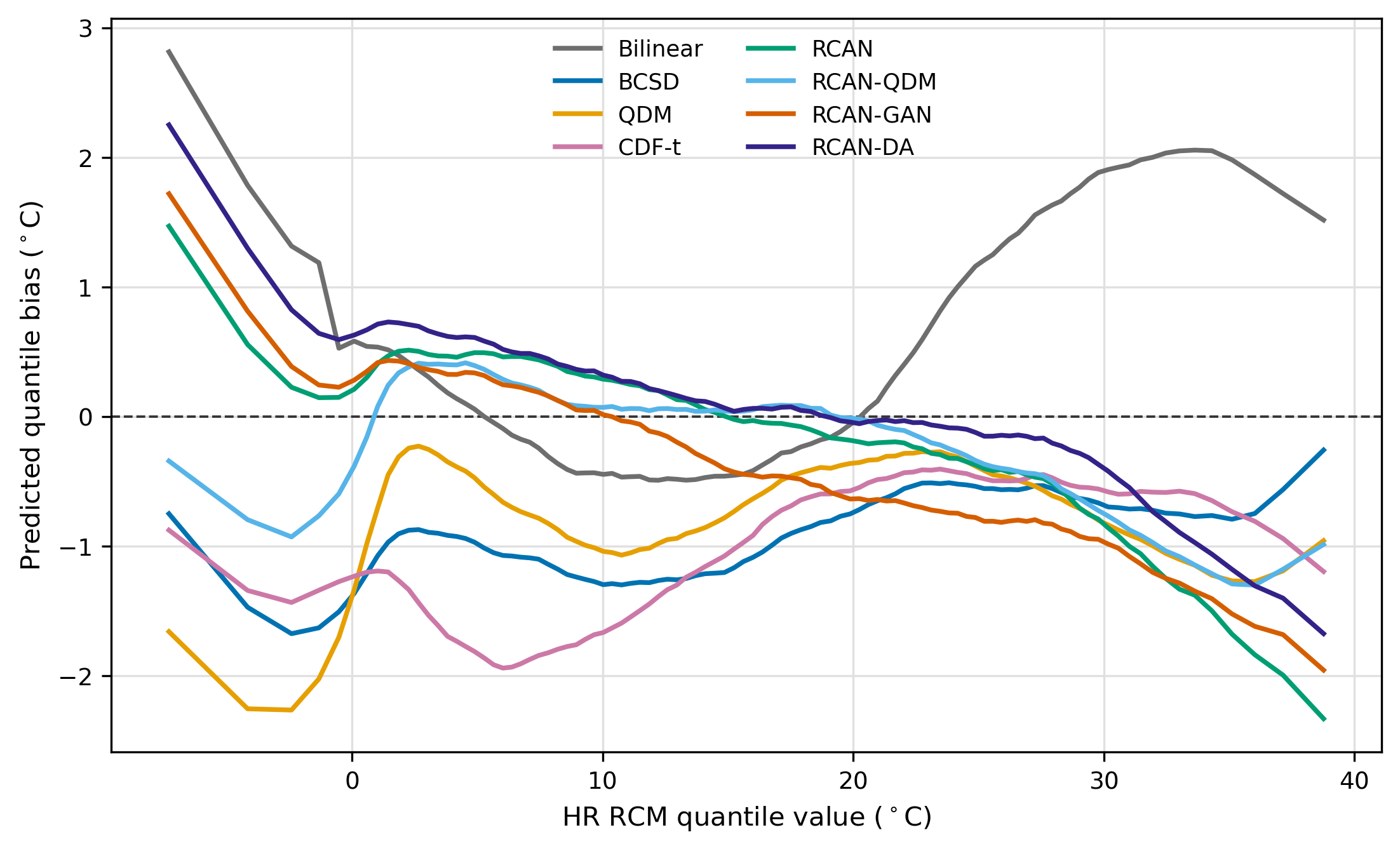}
\caption{Quantile-dependent bias in downscaled daily near-surface air temperature for the CanESM2-RCA4 dataset during 2071--2099. The horizontal axis gives the HR RCM temperature value at each quantile from the 1st to the 99th percentile, and the vertical axis shows the corresponding predicted quantile minus the HR RCM quantile. Positive values indicate warm bias, negative values indicate cold bias, and the dashed zero line represents unbiased reproduction of the target temperature distribution.}
\label{Fig:fig_quantile_bias_curves}
\end{figure}

In this section, we examine quantile-dependent bias and the spatial bias of upper-tail temperature to determine how well each method reproduces both typical and extreme conditions. Figure \ref{Fig:fig_quantile_bias_curves} examines how model bias varies across the temperature distribution. Most methods reproduce intermediate temperatures more accurately than the distribution tails, with biases generally approaching zero around 15--20\,\textdegree C. Bilinear interpolation shows large positive biases at both tails. The statistical methods generally produce negative biases over much of the distribution, although their behavior differs at the coldest quantiles. The DL models show a common pattern in which the coldest quantiles are predicted too warm, whereas the warmest quantiles are predicted too cool, which indicates some compression of the predicted distribution. Domain adaptation maintains relatively small bias over a broad range of intermediate temperatures but still underestimates the upper tail, showing that improved overall downscaling accuracy does not eliminate distributional bias.

\begin{figure}[htbp]
    \centering
    \includegraphics[scale=0.4]{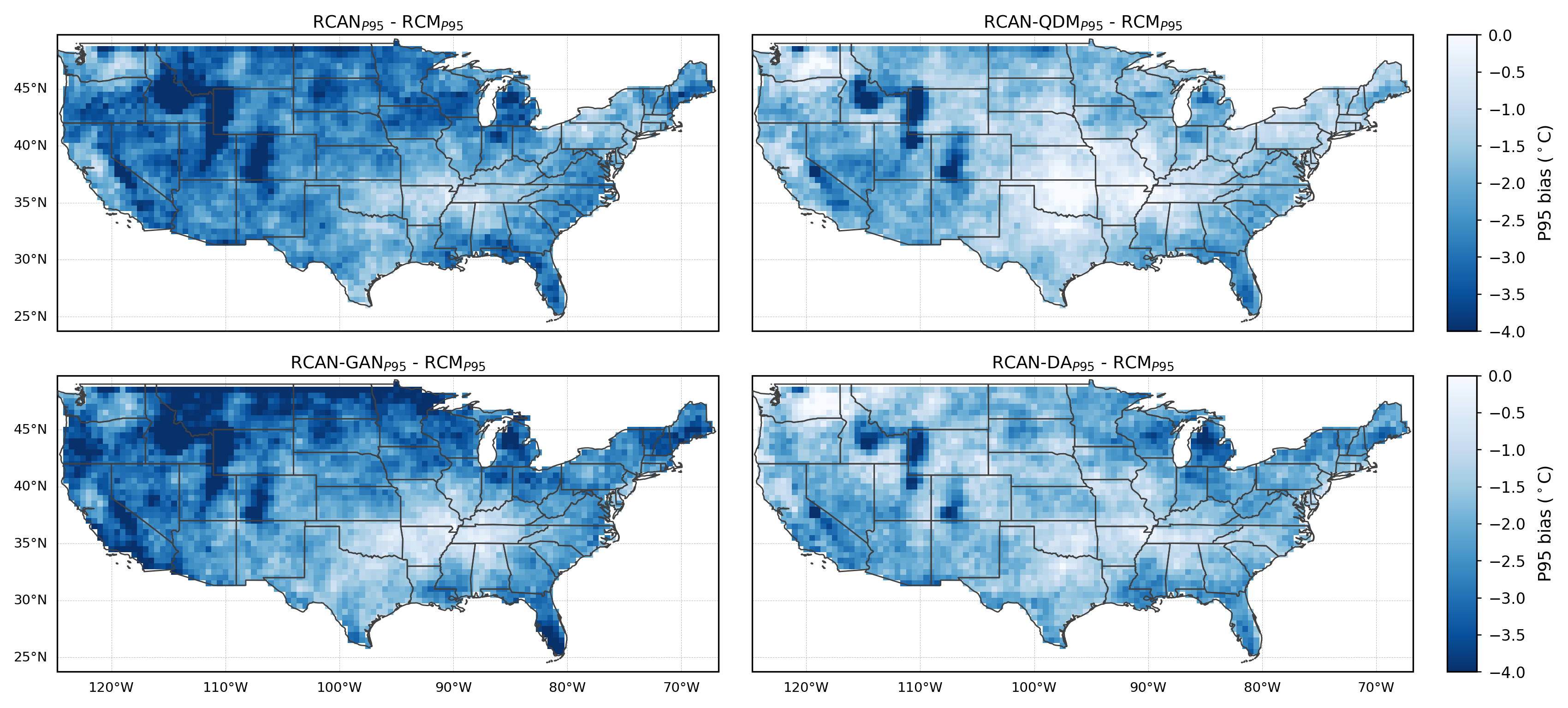}
\caption{Spatial bias in the 95th percentile (P95) of daily near-surface air temperature for the four RCAN-based models using the CanESM2-RCA4 dataset during 2071--2099. At each CONUS grid cell, bias is calculated as the model P95 minus the corresponding HR RCM P95. Darker blue indicates stronger underestimation of high-temperature values, while lighter colors indicate smaller P95 bias.}
\label{Fig:fig_spatial_p95_bias_maps}
\end{figure}

Figure \ref{Fig:fig_spatial_p95_bias_maps} further evaluates how well the DL models reproduce the spatial distribution of extreme high temperatures. All four DL models show negative 95th-percentile biases across much of CONUS, indicating systematic underestimation of high-temperature values. These negative biases are especially evident over parts of the western mountainous region, although their magnitude and spatial extent differ among models. RCAN-GAN and the standard RCAN show relatively widespread negative biases, while RCAN-QDM and RCAN-DA produce smaller biases over portions of the central and eastern United States, indicating that domain adaptation is more effective for extreme-temperature downscaling than the non-adaptive model. However, the 95th-percentile test samples may follow an extreme-tail distribution that differs substantially from the distributions represented in the training and validation sets. Because the domain-adaptation branch aligns source and target features using the available non-extreme samples, it may not have prior information about this tail-specific distribution and therefore cannot fully correct biases in unseen extremes. These results suggest that prediction of temperature extremes remains challenging, even when domain adaptation improves average reconstruction performance and temporal shift.

\section{Discussion}

This study uses domain adaptation to address temporal distribution shift in downscaling under a changing climate. All experiments are conducted within climate model simulations. The RCM outputs are treated as synthetic HR reference fields, forming a pseudo-reality experiment \cite{balmaceda2024use,hernanz2022evaluation}. Future studies should also evaluate the approach using gridded observational or reanalysis products. However, such applications should carefully quantify the magnitude and nature of the source-target shift. When the shift is very small, as may occur in short observational records, the source-trained model may already generalize adequately, leaving little discrepancy for domain adaptation to correct. Conversely, when the shift is too large or the predictor-target relationship changes substantially, source and target supports may have insufficient overlap, and alignment can lead to negative transfer \cite{wang2019characterizing}. Domain adaptation is therefore most likely to be useful for an intermediate, identifiable shift that is large enough to affect model performance but not so large that the historical relationship becomes irrelevant.

The framework could also be extended from temporal to spatial adaptation. In this setting, a data-abundant region with extensive HR data coverage would serve as the labeled source domain, while a data-scarce region would serve as the target domain. Historical LR-HR pairs from the source region could be combined with unlabeled coarse fields and geographic information from the target region to learn representations that transfer across space. For climate downscaling, spatial adaptation could support applications in regions where observations are sparse or dynamical downscaling products are unavailable.

Another extension is application to variables other than temperature, particularly precipitation. Precipitation downscaling is more challenging because precipitation is intermittent, highly skewed, spatially heterogeneous, and strongly controlled by unresolved sub-grid-scale physical processes \cite{maraun2010precipitation,prein2015review,bador2020impact}. We initially considered precipitation in this study, but the historical-future precipitation distribution shift in the selected GCM-RCM simulations was relatively weak. Under this condition, the domain classifier receives a weak separation signal, and any improvement from domain adaptation would be difficult to distinguish from ordinary model variability. We therefore restricted the present analysis to temperature, for which the temporal shift is clear and progressively increases across the validation periods. However, this does not imply that precipitation or other variables are inherently unsuitable for temporal domain adaptation. Future work could examine other datasets, emission scenarios, or regions, where the shift is stronger, and may require more complex model architectures rather than directly reusing the temperature configuration.

\section{Conclusion}

This study examine temporal OOD shift in DL-based climate downscaling under future warming conditions. Using daily near-surface temperature over CONUS from GCM-RCM simulations, we evaluate whether models trained on historical LR-HR pairs can remain reliable when applied to progressively warmer future periods. The validation results confirm that future climate states introduce a clear temporal distribution shift, making downscaling more difficult as the target period moves farther from the historical training regime. To address this challenge, we propose a domain-adaptive downscaling framework that combines supervised HR reconstruction on historical data with temporal domain alignment between historical and future LR inputs. Across three future validation periods and multiple GCM-RCM combinations, the domain-adaptive model consistently outperforms bilinear interpolation, statistical bias-correction methods, the non-adaptive DL model, and other approaches previously shown to be useful under non-stationary climate conditions. The improvements are most evident in the end-of-century period, suggesting that temporal domain adaptation becomes especially valuable as the climate shift strengthens. Spatial analyses further show that RCAN-DA improves reconstruction over topographically complex and high-elevation regions, where coarse GCM inputs provide limited information about fine-scale temperature gradients. The model also achieves stronger spatiotemporal correlation with the HR RCM target, indicating better preservation of regional temperature structure. In the bias and extreme-temperature analyses, temporal domain adaptation improves 95th-percentile temperature performance relative to the non-adaptive model, although some underestimation of variability and negative upper-tail bias remain.

Overall, this work demonstrates that framing future climate downscaling as a domain adaptation problem can improve the robustness of DL downscaling models under non-stationary climate conditions. By aligning historical and future representations during training, the proposed framework improves temporal transferability, strengthens spatial consistency, and provides a more reliable pathway for applying downscaling models to future climate projections.

\section{Acknowledgments}
This study was funded by the United States Department of Defense | Strategic Environmental Research and Development Program (SERDP) - RC20-1183 and the Monsoon Mission III project from the Ministry of Earth Sciences of the Government of India through the Indian Institute of Tropical Meteorology.

\section{Data Availability}
The GCM outputs can be downloaded at  \url{https://metagrid.esgf-west.org/search}. The CORDEX RCM outputs can be downloaded at \url{https://gdex.ucar.edu/datasets/d316009/dataaccess/#}. PRISM elevation data can be downloaded at \href{https://prism.oregonstate.edu/}{here}. The codes used for data preprocessing, training and validation of the downscaling models can be found at \url{https://github.com/shuochenw/downscale}.

\bibliography{tackling_climate_workshop}

\newpage
\appendix
\section{Appendix}
\renewcommand{\thefigure}{A\arabic{figure}}
\setcounter{figure}{0}
\renewcommand{\thetable}{A\arabic{table}}
\setcounter{table}{0}

\begin{table}[htbp]
\centering
\caption{Daily near-surface air temperature statistics for the HR RCM target fields. Each RCM simulation is identified by its driving GCM-RCM pair. Mean shift is computed relative to the historical training period within the same RCM target dataset. The unit is °C.}
\label{tab:temperature_distribution_summary}
\begin{tabular}{llrrrr}
\toprule
Dataset (GCM-RCM) & Period & Mean & Std. & Median & Mean shift \\
\midrule
\multirow{4}{*}{CanESM2-RCA4}
& Training & 10.6439 & 11.5841 & 11.0845 & 0.0000 \\
& 2006--2040 & 12.5943 & 11.4213 & 12.9081 & 1.9504 \\
& 2041--2070 & 14.5418 & 11.3556 & 14.6472 & 3.8979 \\
& 2071--2099 & 16.6485 & 11.3301 & 16.6464 & 6.0047 \\
\midrule
\multirow{4}{*}{CanESM2-CanRCM4}
& Training & 11.3934 & 11.3206 & 11.8346 & 0.0000 \\
& 2006--2040 & 13.1415 & 11.3656 & 13.5002 & 1.7481 \\
& 2041--2070 & 14.7497 & 11.4050 & 15.0226 & 3.3563 \\
& 2071--2099 & 16.6504 & 11.4066 & 17.0810 & 5.2570 \\
\midrule
\multirow{4}{*}{EC-EARTH-RCA4}
& Training & 9.6773 & 10.9556 & 10.2193 & 0.0000 \\
& 2006--2040 & 10.9600 & 10.9548 & 11.4213 & 1.2827 \\
& 2041--2070 & 12.6383 & 10.6044 & 12.9690 & 2.9611 \\
& 2071--2099 & 14.3552 & 10.4994 & 14.5153 & 4.6779 \\
\bottomrule
\end{tabular}
\end{table}

\begin{table}[htbp]
\centering
\caption{Downscaling performance for daily near-surface air temperature over CONUS using the CanESM2-RCA4 paired simulation for other backbone networks.}
\label{tab:backbones}
\begin{tabular}{llrrrr}
\hline
Period & Backbone & MSE & RMSE & PSNR & SSIM \\
\hline
\multirow{4}{*}{2006--2040}
& SRResNet & 5.107 & 2.260 & 31.490 & 0.87818 \\
& EDSR & 5.368 & 2.317 & 31.273 & 0.87533 \\
& RRDB & 4.809 & 2.193 & 31.751 & 0.88133 \\
& UNet & 4.883 & 2.210 & 31.684 & 0.87966 \\
\hline
\multirow{4}{*}{2041--2070}
& SRResNet & 5.085 & 2.255 & 31.246 & 0.88398 \\
& EDSR & 5.554 & 2.357 & 30.862 & 0.88570 \\
& RRDB & 4.864 & 2.205 & 31.439 & 0.88969 \\
& UNet & 5.067 & 2.251 & 31.261 & 0.88575 \\
\hline
\multirow{4}{*}{2071--2099}
& SRResNet & 5.465 & 2.338 & 30.615 & 0.88828 \\
& EDSR & 6.354 & 2.521 & 29.961 & 0.89388 \\
& RRDB & 5.208 & 2.282 & 30.824 & 0.89638 \\
& UNet & 5.491 & 2.343 & 30.594 & 0.89190 \\
\hline
\end{tabular}
\end{table}

\begin{table}[htbp]
\centering
\caption{Validation-period downscaling performance for daily near-surface air temperature over CONUS using the CanESM2-CanRCM4 simulation. Metrics are computed against the HR RCM target fields for three future validation periods. Bold values indicate the best-performing method for each metric within each validation period.}
\label{tab:canesm2_canrcm4_validation_metrics}
\begin{tabular}{llrrrr}
\hline
Period & Method & Full MSE & Full RMSE & PSNR & SSIM \\
\hline
\multirow{8}{*}{2006--2040}
& Bilinear & 25.204 & 5.020 & 23.840 & 0.65172 \\
& BCSD & 1.955 & 1.398 & 34.943 & 0.92855 \\
& QDM & 2.223 & 1.491 & 34.385 & 0.91928 \\
& CDF-t & 2.631 & 1.622 & 33.653 & 0.91552 \\
& RCAN & 1.746 & 1.322 & 35.433 & 0.92704 \\
& RCAN-QDM & 1.693 & 1.301 & 35.567 & 0.93650 \\
& RCAN-GAN & 1.765 & 1.328 & 35.388 & 0.93171 \\
& RCAN-DA & \textbf{1.536} & \textbf{1.239} & \textbf{35.991} & \textbf{0.94070} \\
\hline
\multirow{8}{*}{2041--2070}
& Bilinear & 28.915 & 5.377 & 22.929 & 0.64419 \\
& BCSD & 2.123 & 1.457 & 34.272 & 0.92902 \\
& QDM & 2.918 & 1.708 & 32.890 & 0.90428 \\
& CDF-t & 2.996 & 1.731 & 32.776 & 0.91360 \\
& RCAN & 1.836 & 1.355 & 34.902 & 0.92788 \\
& RCAN-QDM & 1.771 & 1.331 & 35.059 & 0.93837 \\
& RCAN-GAN & 1.870 & 1.368 & 34.821 & 0.93362 \\
& RCAN-DA & \textbf{1.637} & \textbf{1.279} & \textbf{35.401} & \textbf{0.94107} \\
\hline
\multirow{8}{*}{2071--2099}
& Bilinear & 33.383 & 5.778 & 22.292 & 0.64226 \\
& BCSD & 2.467 & 1.571 & 33.605 & 0.92946 \\
& QDM & 4.356 & 2.087 & 31.136 & 0.88498 \\
& CDF-t & 3.522 & 1.877 & 32.059 & 0.91183 \\
& RCAN & 1.991 & 1.411 & 34.536 & 0.93026 \\
& RCAN-QDM & 1.950 & 1.396 & 34.628 & 0.94055 \\
& RCAN-GAN & 2.040 & 1.428 & 34.432 & 0.93662 \\
& RCAN-DA & \textbf{1.803} & \textbf{1.343} & \textbf{34.967} & \textbf{0.94358} \\
\hline
\end{tabular}
\end{table}

\begin{table}[htbp]
\centering
\caption{Validation-period downscaling performance for daily near-surface air temperature over CONUS using the EC-EARTH-RCA4 simulation. Metrics are computed against the HR RCM target fields for three future validation periods. Bold values indicate the best-performing method for each metric within each validation period.}
\label{tab:ecearth_rca4_validation_metrics}
\begin{tabular}{llrrrr}
\hline
Period & Method & Full MSE & Full RMSE & PSNR & SSIM \\
\hline
\multirow{8}{*}{2006--2040}
& Bilinear & 34.202 & 5.848 & 23.631 & 0.55300 \\
& BCSD & 18.409 & 4.291 & 26.321 & 0.76588 \\
& QDM & 18.567 & 4.309 & 26.284 & 0.76517 \\
& CDF-t & 19.230 & 4.385 & 26.131 & 0.75444 \\
& RCAN & 10.665 & 3.266 & 28.692 & 0.81146 \\
& RCAN-QDM & 17.307 & 4.160 & 26.589 & 0.78102 \\
& RCAN-GAN & 10.580 & 3.253 & 28.727 & 0.81098 \\
& RCAN-DA & \textbf{10.229} & \textbf{3.198} & \textbf{28.873} & \textbf{0.81951} \\
\hline
\multirow{8}{*}{2041--2070}
& Bilinear & 34.759 & 5.896 & 23.568 & 0.56233 \\
& BCSD & 16.946 & 4.117 & 26.688 & 0.78270 \\
& QDM & 17.513 & 4.185 & 26.545 & 0.77679 \\
& CDF-t & 17.869 & 4.227 & 26.458 & 0.76906 \\
& RCAN & 10.447 & 3.232 & 28.789 & 0.82121 \\
& RCAN-QDM & 17.852 & 4.225 & 26.462 & 0.79284 \\
& RCAN-GAN & 10.428 & 3.229 & 28.797 & 0.82122 \\
& RCAN-DA & \textbf{9.787} & \textbf{3.128} & \textbf{29.072} & \textbf{0.83177} \\
\hline
\multirow{8}{*}{2071--2099}
& Bilinear & 36.945 & 6.078 & 22.741 & 0.56479 \\
& BCSD & 16.784 & 4.097 & 26.168 & 0.79502 \\
& QDM & 17.938 & 4.235 & 25.879 & 0.77856 \\
& CDF-t & 17.587 & 4.194 & 25.965 & 0.77977 \\
& RCAN & 10.220 & 3.197 & 28.322 & 0.83597 \\
& RCAN-QDM & 17.770 & 4.215 & 25.920 & 0.80683 \\
& RCAN-GAN & 10.610 & 3.257 & 28.159 & 0.83543 \\
& RCAN-DA & \textbf{9.531} & \textbf{3.087} & \textbf{28.625} & \textbf{0.84209} \\
\hline
\end{tabular}
\end{table}

\begin{figure}[htbp]
    \centering
    \includegraphics[scale=0.33]{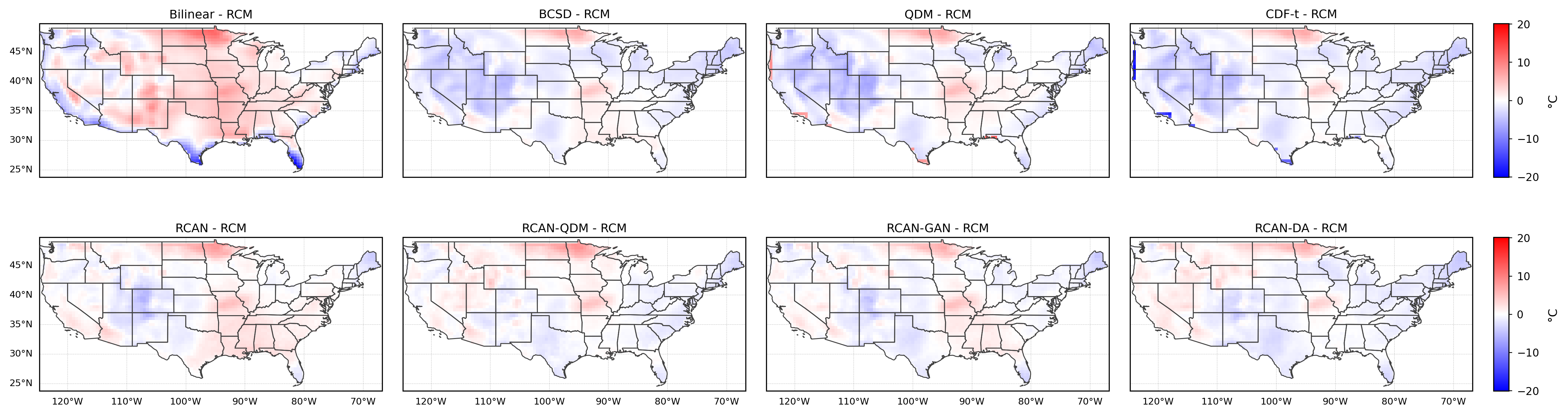}
\caption{Spatial error maps of downscaled daily near-surface air temperature for the CanESM2-RCA4 dataset during the 2006--2040 validation period. Each panel shows the difference between a downscaling method and the HR RCM reference, computed as prediction minus RCM target.}
\label{Fig:prediction_error_maps}
\end{figure}

\begin{figure}[htbp]
    \centering
    \includegraphics[scale=0.33]{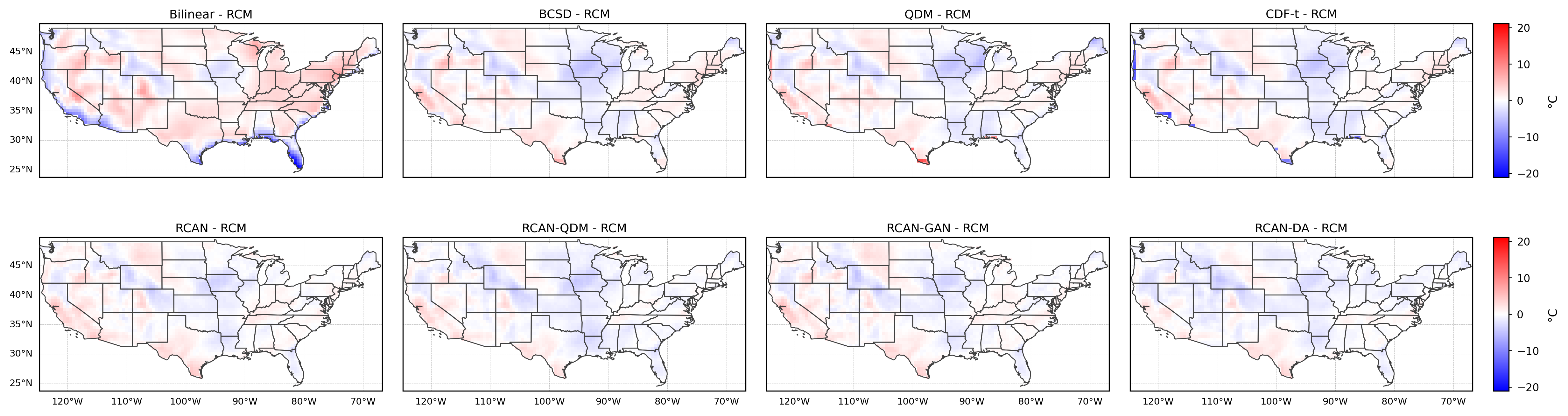}
\caption{Same as Figure \ref{Fig:prediction_error_maps}, but for CanESM2-CanRCM4.}
\label{Fig:prediction_error_maps_CanESM2_CanRCM4}
\end{figure}

\begin{figure}[htbp]
    \centering
    \includegraphics[scale=0.33]{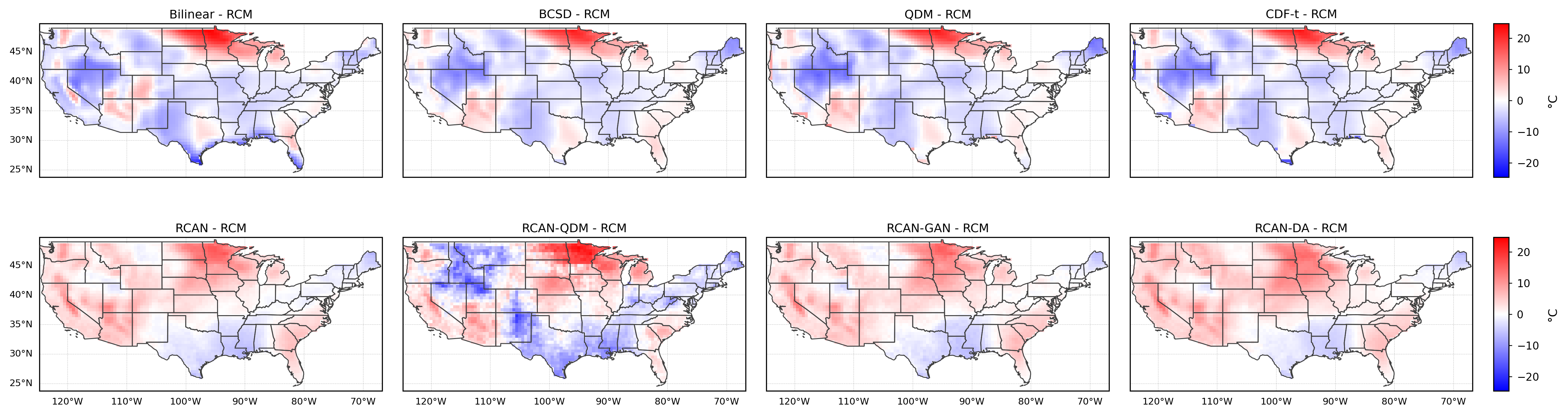}
\caption{Same as Figure \ref{Fig:prediction_error_maps}, but for EC-EARTH-RCA4.}
\label{Fig:prediction_error_maps_EC}
\end{figure}

\begin{figure}[htbp]
    \centering
    \includegraphics[scale=0.4]{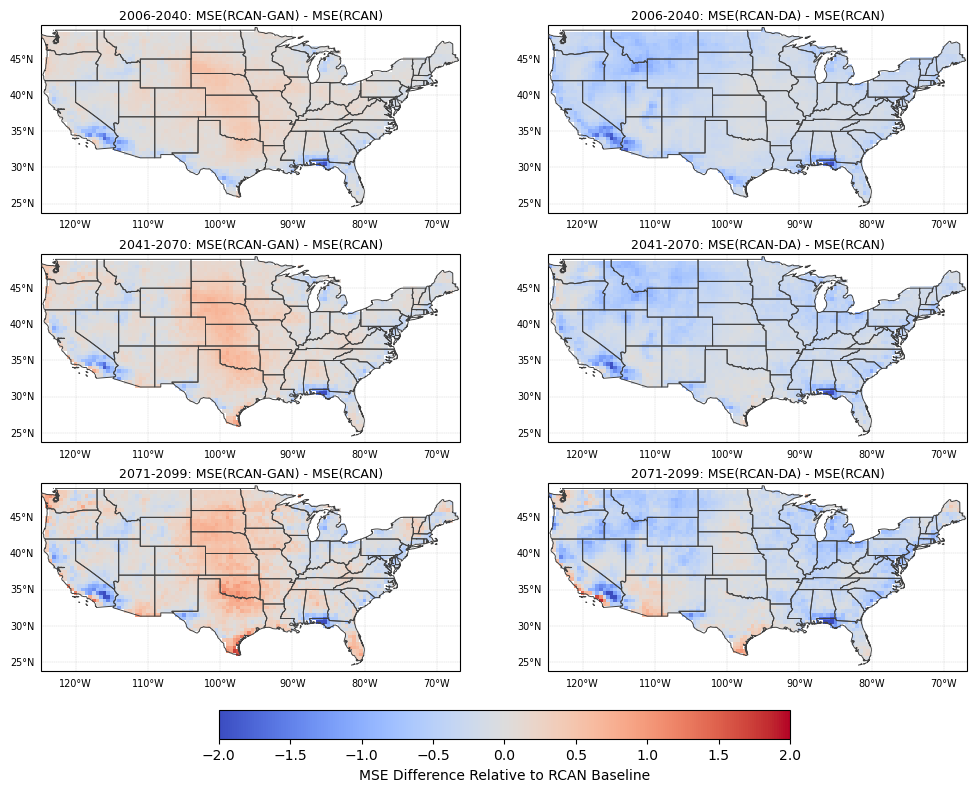}
\caption{Same as Figure \ref{Fig:error_improvement_maps}, but for CanESM2-CanRCM4.}
\label{Fig:error_improvement_maps_CanESM2_CanRCM4}
\end{figure}

\begin{figure}[htbp]
    \centering
    \includegraphics[scale=0.4]{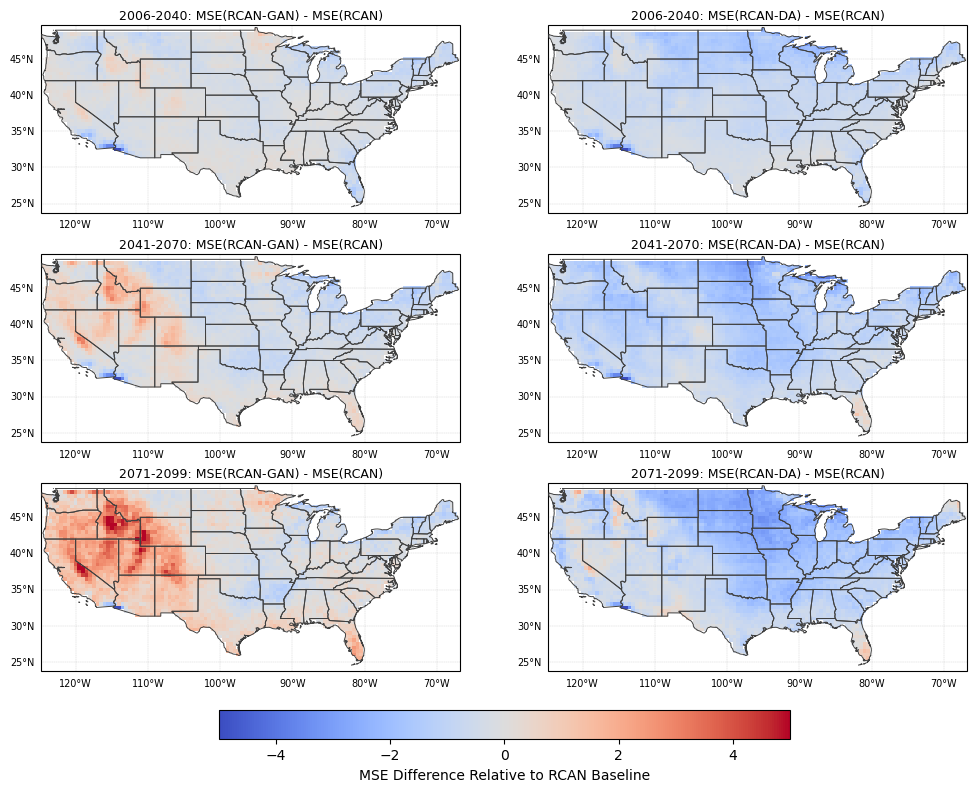}
\caption{Same as Figure \ref{Fig:error_improvement_maps}, but for EC-EARTH-RCA4.}
\label{Fig:error_improvement_maps_EC-EARTH_RCA4}
\end{figure}

\end{document}